\documentclass[letterpaper, 10 pt, conference]{ieeeconf}  

\IEEEoverridecommandlockouts                              

\usepackage{amsmath}
\usepackage{amssymb}
\usepackage{graphicx}
\usepackage{booktabs}

\usepackage{hyperref}

\title{\LARGE \bf
Learning Communication-Conditioned Generative Policies for Decentralized Multi-Agent Collision Avoidance
}

\author{Prajwal Koirala$^{1}$ and Mark Campbell$^{1}$
\thanks{*This work was supported by NSF CPS grant CNS-2211599 and NSF FRR grant IIS-2305532.}
\thanks{$^{1}$Prajwal Koirala and Mark Campbell are with the Sibley School of Mechanical and Aerospace Engineering, Cornell University.
        {\tt\small \{pk596, mc288\}@cornell.edu}}%
\thanks{Code is available at: \url{https://github.com/PrajwalKoirala/CommGenPolicy} .}
}

\begin{document}

\maketitle
\thispagestyle{empty}
\pagestyle{empty}

\begin{abstract}
In this work, we propose a decentralized communication-conditioned generative framework for multi-agent collision avoidance. Agents generate short-horizon action sequences using a flow-matching policy trained from privileged offline demonstrations with access to global state. The demonstrations do not include explicit communication signals; instead, agents learn to exchange and aggregate latent messages that encode interaction-relevant intent under partial observability. This formulation supports flexible inference at test time, where unconditioned generation corresponds to independent behavior and communication-conditioned generation enables coordinated interaction without centralized planning. The resulting policies operate in a fully decentralized manner at execution time, relying only on local observations and learned messages. Combined with a receding-horizon inference scheme, the proposed approach enables efficient single-step inference of short-horizon action sequences and degrades gracefully under communication dropouts. Extensive simulation results demonstrate near-expert collision avoidance performance and strong generalization to denser, unseen multi-agent scenarios, along with zero-shot transfer to real-robot experiments.
\end{abstract}


\section{Introduction}

Multi-agent navigation arises in applications such as autonomous driving, warehouse robotics, aerial swarms, and pedestrian modeling, where agents must reach individual goals while operating in shared environments. As density and heterogeneity increase, ensuring safe and efficient interaction becomes critical. Collision avoidance is central to this problem: each agent must reason about its own dynamics and objectives while anticipating the intentions and future motions of others. The challenge is amplified by decentralization, as agents rely on local observations and must act under partial observability without explicit coordination \cite{oliehoek2016concise}. Designing scalable collision avoidance strategies in such settings remains a fundamental challenge in robotics.

Classical approaches to goal-oriented multi-agent collision avoidance, such as Reciprocal Velocity Obstacles (RVO) and ORCA, rely on strong assumptions about agent dynamics, sensing accuracy, and reciprocal compliance \cite{van2008reciprocal, van2011reciprocal, van2010optimal}. While these methods offer elegant guarantees under idealized conditions, their performance often degrades in dense environments or in the presence of heterogeneous agents whose interaction dynamics deviate from hand-crafted rules. Learning-based approaches provide an appealing alternative by directly capturing implicit interaction patterns from data. Popular learning-based approaches, such as CADRL \cite{everett2021collision} and recent diffusion-based method \cite{shaoul2024multi}, have demonstrated strong performance in multi-agent motion planning. However, many of these methods assume centralized control or access to privileged global state information during execution, limiting their applicability in fully decentralized settings. As a result, learned policies often struggle to generalize to realistic multi-agent systems in which agents must act solely based on local observations.

\begin{figure}
    \centering
    \includegraphics[width=\linewidth]{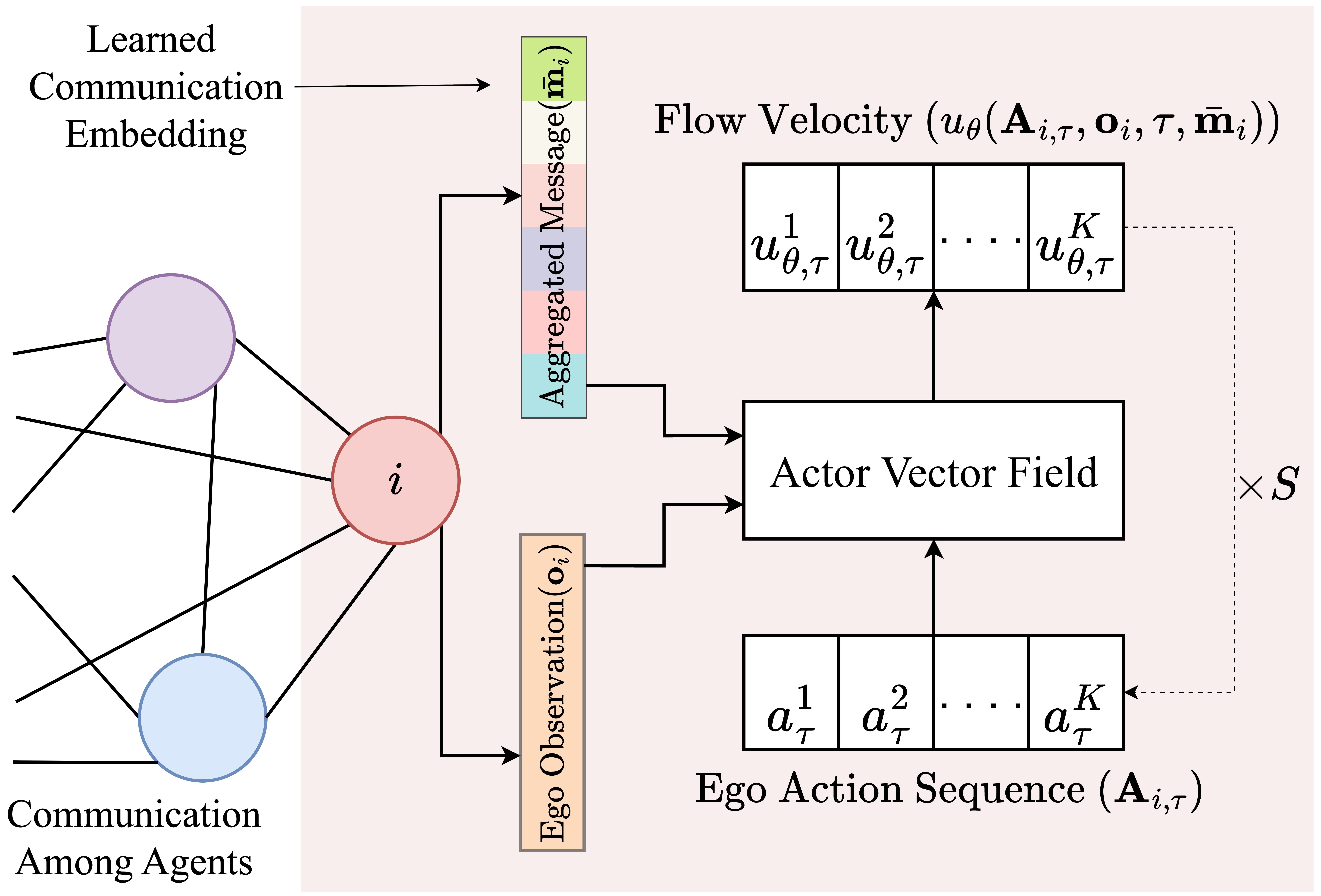}
    \caption{Communication-conditioned generative policy for decentralized multi-agent control.
Each agent aggregates learned inter-agent messages with its local observation to condition a flow-matching policy that generates short-horizon action sequences via iterative refinement. The illustration highlights an ego agent (right), whose action sequence of horizon length $K$ is denoised by repeatedly applying a learned velocity field for $S$ integration steps, while coordination emerges through message exchange without centralized planning.}
    \label{fig:communication-conditioned-introfigure}
    \vspace{-10pt}
\end{figure}

Communication mitigates partial observability and enables coordination in decentralized multi-agent systems. Yet, learning-based communication in robotics and autonomous driving fleets has largely focused on perception-level objectives (e.g., detection and tracking), with limited integration into downstream decision-making and closed-loop evaluation \cite{bai2024survey}. In contrast, control-theoretic approaches provide principled coordination via distributed optimization, often with safety guarantees, but rely on hand-crafted communication protocols and explicit models that restrict representational flexibility and multimodal fusion \cite{filotheou2018decentralized}.
From a learning perspective, communication can be viewed as a representation learning problem: agents must learn what information to transmit, how to aggregate messages from others, and how to use the resulting representation to make interaction-aware decisions \cite{jiang2018learning, guan2024efficient}. How to learn such communication representations end-to-end for decentralized collision avoidance remains an open question.

We argue that generative models (like diffusion \cite{sohl2015deep, ho2020denoising} and flow-matching \cite{lipman2022flow, liu2022flow}) provide a natural and effective framework for communication-conditioned multi-agent decision-making. Deterministic, regression-based policies are often brittle in interactive environments, where multimodal action distributions and distribution shift can lead to compounding errors and cascading failures. In contrast, generative policies can model complex distributions over actions or short-horizon trajectories, allowing them to represent multiple valid interaction strategies, which are essential for safe multi-agent interaction \cite{chi2023diffusion}.  
In this work, we adopt a flow-matching generative policy trained from offline expert demonstrations and condition action generation on learned inter-agent communication. The expert is a privileged policy with access to global state, but the demonstrations themselves do not contain explicit communication signals; instead, communication emerges through end-to-end optimization under partial observability. Thus, our formulation naturally supports flexible communication-conditioning at inference time: unconditioned generation corresponds to agents acting independently based on local observations, while communication-conditioned generation enables interaction-aware decision-making without centralized control. Importantly, the proposed approach also admits efficient single-step inference, addressing a key practical challenge in deploying generative policies for real-time robotic control.

Building on this formulation, we develop a fully decentralized learning and execution framework in which agents rely solely on local observations and learned inter-agent messages at test time. Communication is learned end-to-end and used to condition a receding-horizon generative policy, enabling agents to resolve interactions without centralized planning or explicit coordination signals. In closed-loop simulation, the resulting policies exhibit smooth, interaction-aware behavior and degrade gracefully under communication dropouts, achieving strong collision avoidance performance in dense multi-agent environments. The key contributions of this work are threefold:
\begin{itemize}
    \item a decentralized communication framework that learns interaction-aware messages, aggregating  into a consensus signal for multi-agent collision avoidance;
    \item a communication-conditioned flow-matching policy enabling efficient inference while remaining robust to communication dropouts and interaction complexity;
    \item validation through closed-loop simulation experiments that the proposed approach achieves near-expert collision avoidance performance and robustness to communication dropouts in dense multi-agent environments.
\end{itemize}

\section{Problem Formulation and Preliminaries}
\label{sec:problem}

We consider a decentralized multi-agent collision avoidance problem in a planar workspace. A set of $N$ agents must navigate toward individual goal locations while avoiding collisions with other agents. Each agent operates under partial observability and makes decisions based solely on local sensing and learned communication, without access to centralized coordination or global state at execution time. All agents are assumed to be homogeneous in terms of dynamics, observation models, policy architectures, and communication protocols, and may differ only in their instantaneous states, physical sizes, and goals. This homogeneity assumption enables parameter sharing across agents and supports scalability to varying team sizes.

\subsection{Agent Dynamics}
Each agent $i \in \{1,\dots,N\}$ is modeled as a circular robot with unicycle dynamics. Let $\mathbf{p}_i^{(t)} \in \mathbb{R}^2$ and $\theta_i^{(t)} \in \mathbb{R}$ denote the position and heading of agent $i$ at time $t$. The state evolves according to
\begin{equation} \label{unicycle_agent_dynamics}
\begin{aligned}
\mathbf{p}_i^{(t+1)} &= \mathbf{p}_i^{(t)} + \mathbf{v}_i^{(t)} \Delta t, \\
\theta_i^{(t+1)} &= \theta_i^{(t)} + \Delta \theta_i^{(t)}, \\
\mathbf{v}_i^{(t+1)} &= v_i^{\text{cmd},(t)}
\begin{bmatrix}
\cos \theta_i^{(t+1)} \\
\sin \theta_i^{(t+1)}
\end{bmatrix},
\end{aligned}
\end{equation}
where $(v_i^{\text{cmd}}, \Delta \theta_i)$ are the commanded linear speed and heading change issued by the agent’s policy, and $\Delta t$ denotes the control timestep.

\subsection{Observation Model}
\label{sec:observation_model}

At each timestep, agent $i$ receives a local observation vector $\mathbf{o}_i^{(t)}$ expressed in its ego-centric coordinate frame. The observation consists of goal-related information, self-state features, and limited information about a nearby agent:
\begin{equation}
\mathbf{o}_i =
\begin{bmatrix}
d_i^{\text{goal}} &
\phi_i &
v_i^{\text{pref}} &
r_i &
p_{i,1}^{\parallel} &
p_{i,1}^{\perp} &
v_{i,1}^{\parallel} &
v_{i,1}^{\perp}
\end{bmatrix}^\top .
\end{equation}
Here, $d_i^{\text{goal}} = \|\mathbf{g}_i - \mathbf{p}_i\|$ denotes the distance to the goal, $\phi_i$ is the relative heading toward the goal in the ego frame, $v_i^{\text{pref}}$ is the preferred speed, and $r_i$ is the ego agent’s radius.

The remaining terms describe the relative position and velocity of the nearest neighboring agent, projected onto the ego-frame basis $\{\mathbf{e}_i^{\parallel}, \mathbf{e}_i^{\perp}\}$. Restricting the observation to the closest neighbor ensures scalability and reflects realistic sensing limitations, while capturing the most safety-critical interaction in dense environments.

\subsection{Action Space and Receding-Horizon Control}

Each agent policy operates over a continuous per-timestep control action space
\begin{equation}
\mathbf{a}_i =
\begin{bmatrix}
v_i &
\Delta \theta_i
\end{bmatrix}^\top \in \mathbb{R}^2,
\end{equation}
corresponding to a speed adjustment and heading change.
At any timestep $t$, rather than predicting a single action, the proposed policy produces a short-horizon sequence of actions at each timestep:
\begin{equation}
\mathbf{A}_i^{(t)} =
\begin{bmatrix}
\mathbf{a}_i^{(t,1)} &
\mathbf{a}_i^{(t,2)} &
\cdots &
\mathbf{a}_i^{(t,K)}
\end{bmatrix}^\top \in \mathbb{R}^{K \times 2}.
\end{equation}
Only the first action $\mathbf{a}_i^{(t,1)}$ is executed, and the process is repeated at the next timestep, resulting in a receding-horizon control scheme \cite{terzi2018learning}. This formulation, detailed in section \ref{sec:inference}, enables short-term planning while maintaining reactivity to changes in the environment.

\subsection{Decentralized Communication}

To mitigate partial observability, agents are allowed to exchange learned communication messages. At each timestep, agent $i$ computes a latent message $\mathbf{m}_i \in \mathbb{R}^{d_m}$ based solely on its local observation. Messages received from other agents are aggregated into a fixed-dimensional communication embeddings, which conditions the ego agent’s policy. Communication is fully decentralized: agents do not have access to global state, agent identities, or centralized coordination at execution time. The specific message generation, aggregation mechanism, and conditioning strategy are detailed in Section~\ref{sec:method}.

\subsection{Task Objective and Data Collection}

The objective of each agent is to reach its goal location while avoiding collisions with other agents. A collision is defined to occur when the distance between any pair of agents $i \neq j$ satisfies
$\|\mathbf{p}_i - \mathbf{p}_j\| \leq r_i + r_j$.
An episode terminates when all agents reach their goals within a tolerance $\varepsilon_{\text{goal}}$, a collision occurs, or a maximum time horizon $T_{\max}$ is exceeded. 

Training data is collected using the CADRL \cite{everett2021collision}, which serves as a privileged expert policy with access to the states of all other agents in the scene, expressed in the ego agent’s local reference frame. This observation space is strictly richer than the local observation model described in Section~\ref{sec:observation_model}. Although CADRL demonstrations do not include explicit communication signals, they exhibit coordinated collision-avoidance behaviors enabled by access to centralized multi-agent state information. 
While learning-based methods achieve strong performance, their reliance on privileged observations during execution limits applicability in fully decentralized settings. In contrast, our goal is to learn policies that operate under decentralized execution with partial observability and uncertain communication. To this end, we leverage privileged expert demonstrations during training and distill the resulting behaviors into agent policies that rely solely on local observations and learned inter-agent communication for robust collision avoidance \cite{li2021learning}.

\subsection{Generative Models and Flow Matching Policies}

Generative models provide a flexible alternative to unimodal policies by modeling distributions over actions or short-horizon trajectories. In robotics, diffusion-based policies generate action sequences through iterative denoising and have demonstrated strong performance when trained from offline demonstrations. These approaches interpret control as sampling from a learned conditional distribution rather than predicting a single action \cite{chi2023diffusion}. Flow matching offers a closely related but deterministic and computationally efficient alternative. Instead of learning a reverse-time stochastic process, flow matching directly learns a time-dependent vector field that transports samples from a simple base distribution to the data distribution along a predefined interpolation \cite{lipman2022flow, liu2022flow}. Let $\mathbf{x}_\tau \in \mathbb{R}^d$ denote an intermediate latent sample at interpolation time $\tau \in [0,1]$, where $\tau=0$ corresponds to noise (e.g., $\mathbf{x}_0 \sim \mathcal{N}(0, I)$) and $\tau=1$ corresponds to dataset sample $\mathbf{x}_1 \sim \mathcal{D}$ (eg. action sequences). Flow matching learns a velocity field $u_\theta$ satisfying
\begin{equation}
    \frac{d \mathbf{x}_\tau}{d \tau} = u_\theta(\mathbf{x}_\tau, \tau, o),
\end{equation}
where $o$ denotes the conditioning input, such as the agent’s observation. Given a learned velocity field $u_\theta$, samples from the data distribution can be generated by numerically integrating this ordinary differential equation, e.g., using Euler updates. As detailed in Section~\ref{sec:generative_policy}, the conditioning input may additionally include learned inter-agent communication signals, enabling interaction-aware action generation while preserving decentralized execution.

\section{Methodology}
\label{sec:method}

\subsection{Communication and Message Passing}
\label{sec:communication}

In our decentralized multi-agent setting, each agent acts based on its local observation and information exchanged through learned communication signals. This section describes the communication mechanism used to summarize interaction-relevant information from other agents and provide a conditioning signal for the generative policy introduced in Section~\ref{sec:generative_policy}. We first define how agents emit messages from local observations, and then describe how these messages are aggregated into a shared representation that supports coordinated decision-making under decentralized execution.

\paragraph{Message Emission.}
At each time step \(t\), agent \(i\) produces a latent message
\begin{equation}
\mathbf{m}_i^{(t)} = \mathcal{M}_{\theta_1}\big(\mathbf{o}_i^{(t)}\big),
\end{equation}
where \(o_i^{(t)}\) denotes the agent’s local observation and \(\mathcal{M}_{\theta_1}\) is a shared message encoder applied independently across agents. The message \(\mathbf{m}_i^{(t)} \in \mathbb{R}^d\) is a compressed latent representation and is not required to explicitly encode physical state or future actions.

\paragraph{Pairwise Interaction Encoding.}
To model directed interactions between agents, each ordered agent pair \((i,j)\) forms a pairwise interaction token
\begin{equation}
\mathbf{z}_{ij}^{(t)} = \mathcal{I}_{\theta_2}\!\left(\mathbf{m}_i^{(t)}, \mathbf{m}_j^{(t)}\right),
\end{equation}
where \(\mathcal{I}_{\theta_2}\) is a learned interaction encoder shared across all agent pairs. In parallel, a scalar compatibility score
\begin{equation}
\alpha_{ij}^{(t)} = \alpha_{\theta_3}\!\left(\mathbf{m}_i^{(t)}, \mathbf{m}_j^{(t)}\right)
\end{equation}
is computed for the same pair. Self-interactions are masked.

\paragraph{Attention-Based Aggregation}
The compatibility scores are normalized across agents to form a stochastic attention \cite{velivckovic2017graph} distribution
\begin{equation}
w_{ij}^{(t)} =
\frac{\exp(\alpha_{ij}^{(t)})}{\sum_{\ell \neq i} \exp(\alpha_{i\ell}^{(t)})},
\end{equation}
and the aggregated communication vector processed by agent \(i\) is given by
\begin{equation}
\bar{\mathbf{m}}_i^{(t)} = \sum_{j \neq i} w_{ij}^{(t)} \, \mathbf{z}_{ij}^{(t)}.
\end{equation}
This aggregation is permutation-invariant and relies only on locally exchanged messages, preserving decentralization. Additionally, the aggregated vector \(\bar{\mathbf{m}}_i^{(t)}\) can be interpreted as a \emph{soft consensus} over interaction-relevant information from other agents, providing a latent summary of how their behavior is expected to influence agent \(i\) over a short horizon. Rather than enforcing explicit agreement or shared plans, the attention-weighted aggregation induces a soft form of consensus by allowing each agent to adapt its internal representation based on a learned, weighted summary of others’ messages. From this perspective, \(\mathbf{m}_i^{(t)}\) characterizes interaction \emph{intent} in a latent and task-dependent manner, without requiring predefined semantics or explicit coordination protocols.

The communication module is intentionally lightweight and free of privileged information. Its parameters are trained jointly with the conditional generative policy and are further shaped by an auxiliary distillation objective \cite{hinton2015distilling, li2021learning} described in Section~\ref{sec:teacher_distillation}. We next describe how the aggregated intent embeddings condition the generation of short-horizon action sequences.

\subsection{Conditional Generative Policy}
\label{sec:generative_policy}

We model each agent’s decision-making process as conditional generation of a short-horizon action sequence, where conditioning occurs through the learned communication embeddings described in Section~\ref{sec:communication}. This separation allows interaction reasoning to be captured in a compact latent form, while action generation is handled by a scalable generative policy.

\paragraph{Flow-Based Policy Parameterization}
We parameterize the policy using a conditional flow-matching model that transports samples from a simple base distribution to the target action-sequence distribution. Let
\(
\boldsymbol{\epsilon} \sim \mathcal{N}(\mathbf{0}, \mathbf{I})
\)
denote a noise sample of the same dimension as the action sequence \(\mathbf{A}_i^{(t)}\). For an interpolation variable \(\tau \sim \mathcal{U}[0,1]\), we define the linear interpolation
\begin{equation}
\mathbf{A}_{\tau} =
\tau \, \mathbf{A}_i^{(t)} + (1 - \tau)\, \boldsymbol{\epsilon}.
\end{equation}
The conditional velocity field
\begin{equation}
u_\theta\!\left(
\mathbf{A}_{\tau}, \mathbf{o}_i^{(t)}, \tau, \bar{\mathbf{m}}_i^{(t)}
\right)
\end{equation}
is implemented as a neural network and predicts the transport velocity at interpolation time \(\tau\), conditioned on the agent’s local observation and aggregated communication embedding.

\paragraph{Conditional Flow-Matching Objective}
The conditional flow-matching loss is defined as
\begin{equation}
\mathcal{L}_{\mathrm{CFM}} =
\mathbb{E}_{\mathbf{A}, \boldsymbol{\epsilon}, \tau}
\left[
\left\|
u_\theta\!\left(
\mathbf{A}_{\tau}, o_i^{(t)}, \tau, \bar{\mathbf{m}}_i^{(t)}
\right)
-
\left(\mathbf{A}_i^{(t)} - \boldsymbol{\epsilon}\right)
\right\|_2^2
\right],
\end{equation}
which trains the model to recover the optimal transport vector field under conditioning. Gradients from this objective propagate through both the generative policy and the communication module, enabling end-to-end learning of interaction-aware action generation.

\paragraph{Unconditional Flow-Matching Objective}
In addition to conditional generation, we incorporate an unconditional flow-matching objective that excludes communication conditioning:
\begin{equation}
\mathcal{L}_{\mathrm{UFM}} =
\mathbb{E}_{\mathbf{A}, \boldsymbol{\epsilon}, \tau}
\left[
\left\|
u_\theta\!\left(
\mathbf{A}_{\tau}, o_i^{(t)}, \tau, \varnothing
\right)
-
\left(\mathbf{A}_i^{(t)} - \boldsymbol{\epsilon}\right)
\right\|_2^2
\right],
\end{equation}
where $\varnothing$ represents the \textit{unconditional embedding}. This objective regularizes the base generative dynamics and stabilizes training, but does not contribute gradients to the communication module. As a result, learned messages are shaped exclusively by interaction-relevant conditional generation rather than by marginal action statistics.

\subsection{Privileged Teacher and Distillation}
\label{sec:teacher_distillation}

While communication in Section~\ref{sec:communication} is learned end-to-end through the conditional generative objective, we additionally introduce a privileged interaction model to shape the semantics of the learned messages. This model acts as a teacher that encodes predictive beliefs about near-term system evolution using information unavailable at test time, and supervises the communication module through an auxiliary distillation objective.

\begin{figure}[b]
    \centering
    \includegraphics[width=\linewidth]{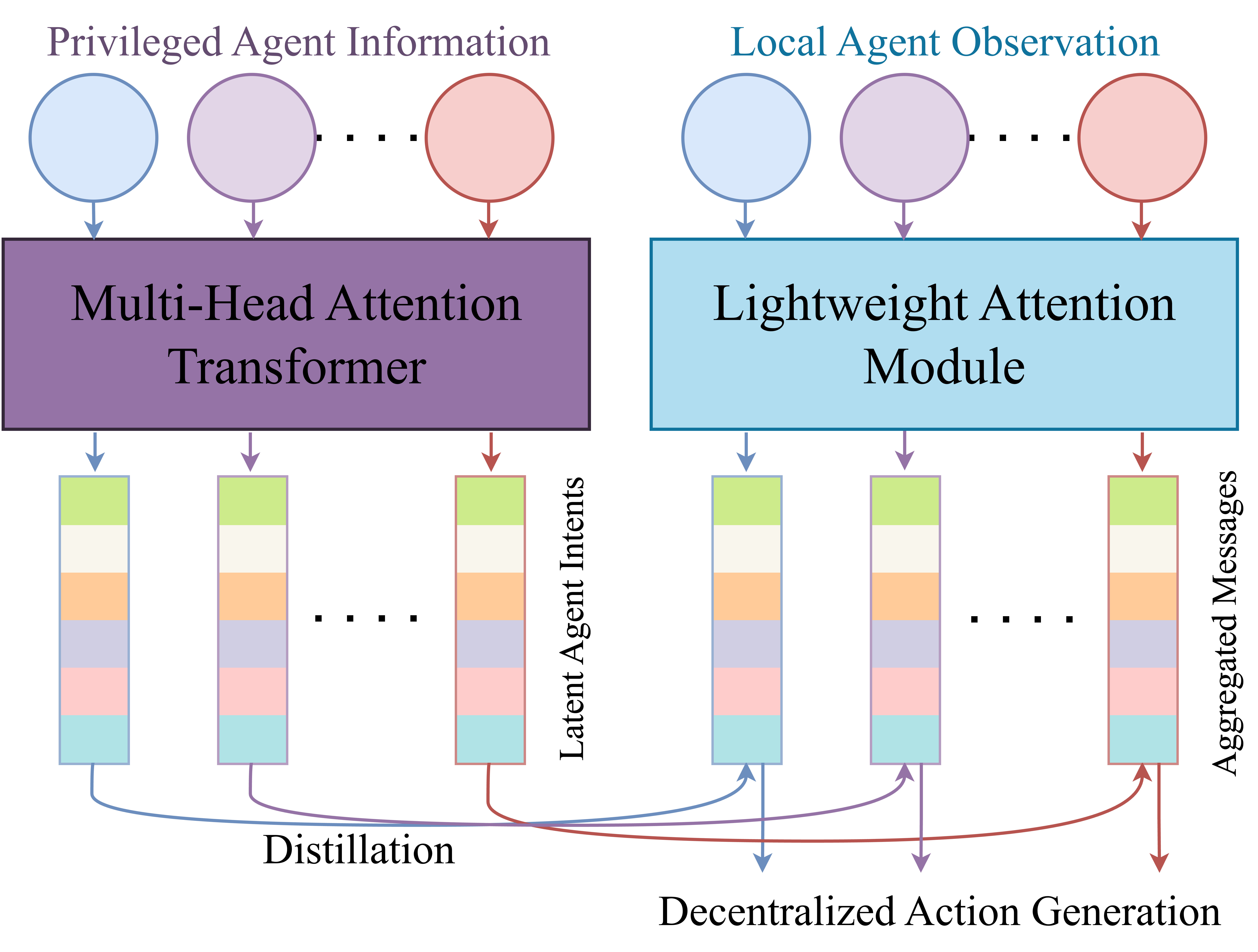}
    \vspace{-20pt}
    \caption{
    Privileged teacher and decentralized distillation. Global interaction structure is learned by a transformer teacher and distilled into a lightweight communication module whose aggregated messages condition decentralized generative policies at test time.
    }
    \label{fig:distillation-introfigure}
\end{figure}

\paragraph{Privileged Interaction Transformer}
We instantiate the teacher as a transformer-based interaction encoder
\begin{equation}
\mathbf{h}^{(t)} = \mathcal{T}_{\phi}\!\left(
\left\{ \left(\mathbf{o}_i^{(t)}, \mathbf{g}_i^{(t)} \right) \right\}_{i=1}^N
\right),
\end{equation}
where \(\mathbf{o}_i^{(t)}\) denotes the local observation of agent \(i\), and \(\mathbf{g}^{(t)}\) denotes privileged global reference information available only during training (global ego and goal- coordinates). The output
\(
\mathbf{h}^{(t)} = \{\mathbf{h}_i^{(t)}\}_{i=1}^N
\)
is a set of agent-wise latent intent embeddings. The transformer architecture \cite{vaswani2017attention} enables joint reasoning over agents and produces representations that encode interaction-relevant predictive structure rather than explicit actions.

\paragraph{Teacher Training Objective}
The teacher is trained independently via a reconstruction objective that encourages each latent embedding \(\mathbf{h}_i^{(t)}\) to capture information sufficient to predict the agent’s immediate future observation:
\begin{equation}
\mathcal{L}_{\mathrm{intent}} =
\mathbb{E}
\left[
\left\|
\mathcal{D}_{\psi}\!\left(\mathbf{h}_i^{(t)}\right)
-
\mathbf{o}_i^{(t+1)}
\right\|_2^2
\right],
\end{equation}
where \(\mathcal{D}_{\psi}\) is a lightweight decoder. Gradients from \(\mathcal{L}_{\mathrm{intent}}\) update both the transformer parameters \(\phi\) and the decoder parameters \(\psi\), yielding a latent representation that reflects the teacher’s predictive belief over near-term state evolution.

\paragraph{Distillation into Communication Messages}
To align the learned communication embeddings with the teacher’s predictive structure, we reuse the decoder \(\mathcal{D}_{\psi}\) to define an auxiliary loss on the aggregated messages \(\bar{\mathbf{m}}_i^{(t)}\) produced by the communication module:
\begin{equation}
\mathcal{L}_{\mathrm{aux}} =
\mathbb{E}
\left[
\left\|
\mathcal{D}_{\psi}\!\left(\bar{\mathbf{m}}_i^{(t)}\right)
-
\mathbf{o}_i^{(t+1)}
\right\|_2^2
\right].
\end{equation}
Importantly, during optimization of \(\mathcal{L}_{\mathrm{aux}}\), gradients are propagated only to the communication parameters and \emph{not} to the decoder \(\mathcal{D}_{\psi}\). The decoder therefore serves as a fixed projection that defines the semantic target space for the messages, preventing collapse to trivial encodings.

This training procedure induces a clean separation of functional roles across components. A privileged transformer, with access to global state information during training, is responsible for learning a predictive belief representation that captures short-horizon system evolution and encodes the latent intent of the expert agents used to generate the data \cite{zheng2025intention}. In parallel, a decentralized communication module is trained to approximate this belief representation using only local observations and inter-agent message passing, thereby learning a distributed surrogate of the privileged belief without direct access to global inputs. The resulting messages are then consumed by a communication-conditioned generative policy, which is trained without exposure to any privileged information and relies solely on the learned messages for coordination. Consequently, the messages encode interaction-relevant intent rather than raw observations or explicit action proposals, enabling effective conditioning of the generative policy while strictly preserving decentralized execution at test time. From a probabilistic perspective, the privileged transformer acts as a \textit{teacher} that defines a latent belief over short-term system dynamics, while the communication module learns a constrained approximation of this belief through an auxiliary distillation objective that shapes the geometry of the message space without directly prescribing policy outputs. The final multi-agent behavior thus emerges from the downstream conditional generative objective described in Section~\ref{sec:generative_policy}, rather than being imposed by the distillation process itself.

\subsection{Overall Training Objectives and Optimization}
\label{sec:training}

We train the full system using a composite objective that combines conditional generation, auxiliary regularization, and privileged supervision. The total loss is given by
\begin{equation}
\mathcal{L}_{\mathrm{total}}
=
\mathcal{L}_{\mathrm{CFM}}
+
\lambda_1 \, \mathcal{L}_{\mathrm{UFM}}
+
\lambda_2 \, \mathcal{L}_{\mathrm{aux}}
+
\mathcal{L}_{\mathrm{intent}},
\end{equation}
where \(\lambda_1, \lambda_2 < 1\) weight auxiliary objectives that support, but do not dominate, the primary conditional generative training signal. Overall, \(\mathcal{L}_{\mathrm{CFM}}\) provides the dominant end-to-end learning signal, while auxiliary losses shape representation geometry and stabilize training without introducing privileged information into inference-time components.

\subsection{Decentralized Inference and Receding-Horizon Execution}
\label{sec:inference}

At evaluation time, agents execute policies in a fully decentralized manner, using only local observations and learned communication. No privileged information, centralized coordination, or teacher models are accessed. Rather than predicting a single control input, the policy generates a short-horizon action sequence $\mathbf{A}_i^{(t)}$ 
of which only the first action \(\mathbf{a}_i^{(t,1)}\) is executed by the agent $i$ at timestep $t$. The remaining actions serve as an internal plan refined over time.

At each time step, agents exchange messages and compute aggregated representations \(\bar{\mathbf{m}}_i^{(t)}\) following Section~\ref{sec:communication}. These aggregated messages condition the generative policy during action sequence generation. Specifically, action sequences are generated by integrating the learned velocity field \(u_\theta\) using a small number of explicit Euler steps.

\paragraph{Initialization (first step of an episode)}
When no prior action sequence is available, generation starts from Gaussian noise 
$\mathbf{A}_{i,0} \sim \mathcal{N}(0, I)$.
Starting from \(\tau=0\), we perform \(S\) Euler integration steps:
\begin{equation}
\mathbf{A}_i \leftarrow \mathbf{A}_i + \frac{1}{S}
u_\theta(o_i^{(t)}, \mathbf{A}_i, \tau, \bar{\mathbf{m}}_i^{(t)}),
\quad
\tau \leftarrow \tau + \frac{1}{S}.
\end{equation}

\paragraph{Receding-horizon warm start}
At subsequent time steps, we warm-start generation from the previously predicted action sequence \(\mathbf{A}_i^{(t-1)}\). To allow adaptation to new observations, we partially corrupt the sequence with a Gaussian sample $\boldsymbol{\epsilon} \sim \mathcal{N}(0, I)$:
\begin{equation}
\tilde{\mathbf{A}}_i^{(t)}
=
\alpha \, \mathbf{A}_i^{(t-1)}
+
(1-\alpha)\boldsymbol{\epsilon},
\quad
\alpha = 1 - \frac{1}{S}.
\end{equation}
A single denoising step is then applied at time \(\tau=\alpha\):
\begin{equation}
\mathbf{A}_i^{(t)}
=
\tilde{\mathbf{A}}_i^{(t)}
+
(1-\alpha)
u_\theta(o_i^{(t)}, \tilde{\mathbf{A}}_i^{(t)}, \alpha, \bar{\mathbf{m}}_i^{(t)}).
\end{equation}

\paragraph{Execution}
Each agent executes only the first action \(\mathbf{a}_i^{(t,1)}\), discards it from the plan, and repeats the procedure at the next time step.

This receding-horizon inference scheme achieves temporal consistency through warm-starting while remaining responsive to newly acquired observations and inter-agent communication. By initializing each inference step from the previously selected action sequence, the procedure implicitly encourages commitment to a coherent mode of behavior, thereby reducing abrupt changes in sequential decisions without the need for explicit regularization or smoothing terms. As a result, temporal smoothness emerges naturally from the inference process itself rather than being imposed by auxiliary objectives. Moreover, the warm-started sequence provides a principled fallback mechanism in the presence of communication dropouts: when the aggregated message \(\mathbf{\bar{m}}_t^i\) is unavailable or unreliable, the agent can continue executing a consistent plan based on its prior inference. Importantly, after initialization, the method requires only a single evaluation of the learned actor vector field \(u_\theta\) per agent at each timestep, yielding a computationally efficient inference procedure that is compatible with real-time decentralized control constraints.

\section{Experiments, Results and Discussions}

\subsection{Overall Results and Generalization Across Agent Scales}
\label{sec:results_scaling}

We first evaluate the overall effectiveness of the proposed communication-conditioned generative policy and its ability to generalize beyond the training regime. All models are trained exclusively on trajectories collected from four-agent collision avoidance scenarios, while evaluation is performed on environments containing $N \in \{4,6,8\}$ agents. Importantly, no additional fine-tuning or retraining is performed when increasing the number of agents, making the six- and eight-agent evaluations a strict test of out-of-distribution (OOD) generalization with respect to agent count. Policies are trained in the Gym Collision Avoidance environment within We report success rate (fraction of agents reaching their goal without collision) and collision rate, averaged over complete episodes, including standard deviation across multiple random seeds and environment initializations.

\begin{figure}[!h]
    \centering
    \includegraphics[width=0.9\linewidth]{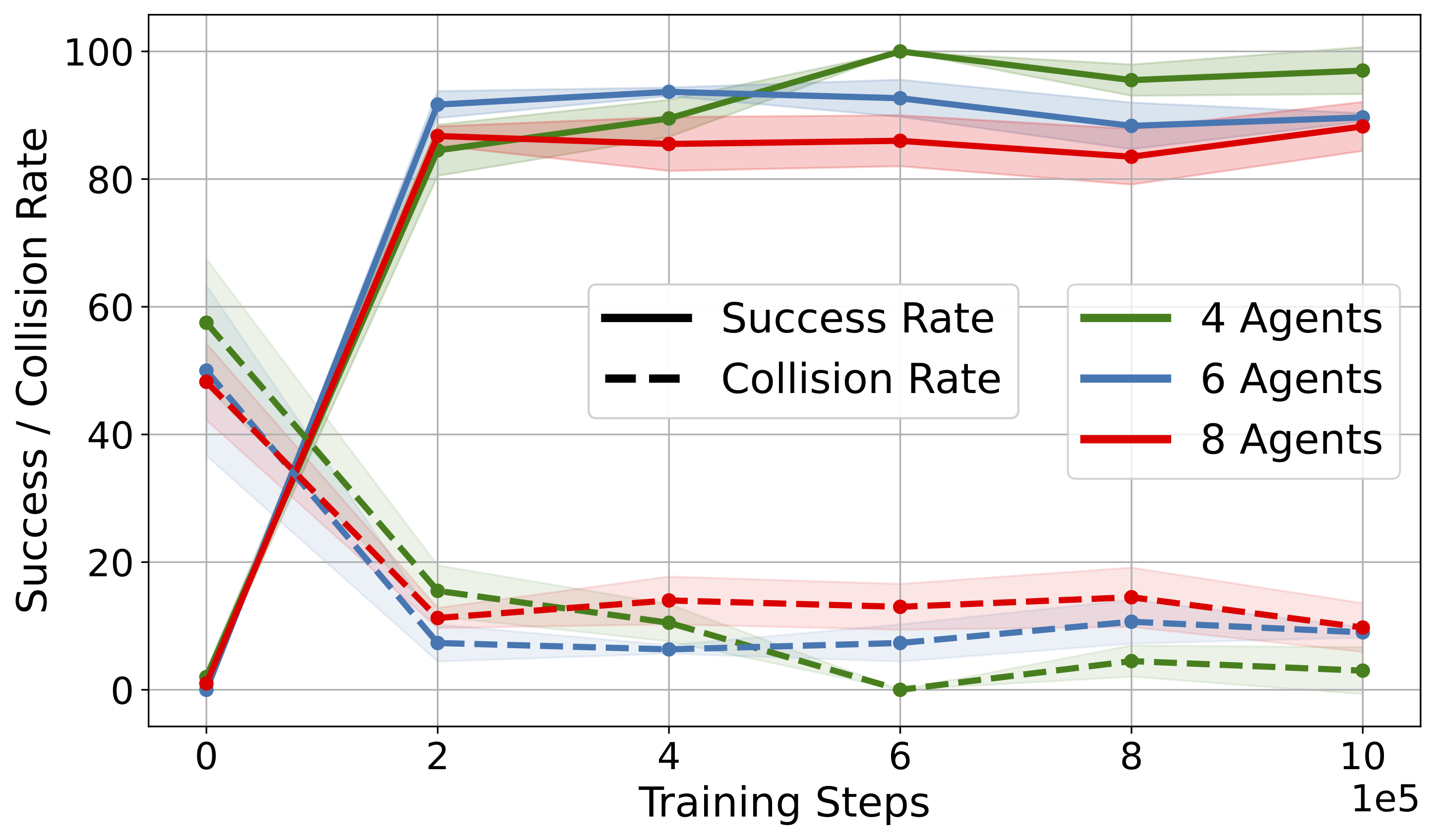}
    \vspace{-5pt}
    \caption{Training dynamics and generalization across agent counts. Success rate and collision rate during training, evaluated periodically on environments with 4, 6, and 8 agents. Policies are trained exclusively on 4-agent demonstrations and evaluated zero-shot on larger agent counts, illustrating stable learning and strong generalization to denser multi-agent scenarios.}
    \label{fig:training-curve-results-figure}
\end{figure}

\paragraph{Learning dynamics}
Figure~\ref{fig:training-curve-results-figure} illustrates the evolution of success and collision rates as the training proceeds. Notably, although optimization is driven solely by four-agent data, the learned policy exhibits an improvement in performance not only on the in-distribution four-agent setting but also on six- and eight-agent scenarios. This indicates that the communication module and flow-matching policy learn interaction patterns that extrapolate to denser agent configurations, rather than overfitting to a fixed agent count.

\begin{table*}[t]
\centering
\caption{Quantitative evaluation of multi-agent collision avoidance under cooperative and partially non cooperative settings.}
\label{tab:main_results}

\resizebox{\textwidth}{!}{
\begin{tabular}{lcccccccc}
\toprule
& \multicolumn{4}{c}{\textbf{4 Agents (Train Scale)}} 
& \multicolumn{4}{c}{\textbf{8 Agents (Generalization)}} \\
\cmidrule(lr){2-5} \cmidrule(lr){6-9}

\textbf{Method}
& \multicolumn{2}{c}{\textbf{Cooperative}}
& \multicolumn{2}{c}{\textbf{Non-Cooperative}}
& \multicolumn{2}{c}{\textbf{Cooperative}}
& \multicolumn{2}{c}{\textbf{Non-Cooperative}} \\
\cmidrule(lr){2-3} \cmidrule(lr){4-5}
\cmidrule(lr){6-7} \cmidrule(lr){8-9}

& Success $\uparrow$ & Collision $\downarrow$
& Success $\uparrow$ & Collision $\downarrow$
& Success $\uparrow$ & Collision $\downarrow$
& Success $\uparrow$ & Collision $\downarrow$ \\
\midrule

\textbf{Proposed} 
& $0.97 \pm 0.02$ & $0.03 \pm 0.02$ & $0.80 \pm 0.01$ & $0.19 \pm 0.02$
& $0.92 \pm 0.02$ & $0.06 \pm 0.02$ & $0.79 \pm 0.02$ & $0.19 \pm 0.02$ \\
\midrule

Proposed w/o Communication 
& $0.31 \pm 0.02$ & $0.69 \pm 0.02$ & $0.36 \pm 0.02$ & $0.64 \pm 0.02$
& $0.52 \pm 0.12$ & $0.47 \pm 0.12$ & $0.48 \pm 0.13$ & $0.50 \pm 0.13$ \\

Proposed w/o Auxiliary Loss 
& $0.86 \pm 0.08$ & $0.14 \pm 0.08$ & $0.75 \pm 0.04$ & $0.25 \pm 0.04$ & $0.93 \pm 0.03$ & $0.07 \pm 0.03$ & $0.77 \pm 0.03$ & $0.23 \pm 0.03$ \\

Proposed w/o Unconditioned Loss 
& $0.87 \pm 0.06$ & $0.13 \pm 0.06$ & $0.81 \pm 0.07$ & $0.19 \pm 0.07$ & $0.88 \pm 0.03$ & $0.11 \pm 0.02$ & $0.81 \pm 0.08$ & $0.17 \pm 0.08$ \\
\midrule

Decentralized Behavioral Cloning (BC) \cite{pomerleau1988alvinn} 
& $0.62 \pm 0.06$ & $0.38 \pm 0.06$ & $0.56 \pm 0.02$ & $0.44 \pm 0.02$ & $0.48 \pm 0.14$ & $0.52 \pm 0.14$ & $0.67 \pm 0.05$ & $0.33 \pm 0.05$ \\

Decentralized BC w/ Attention-Based Communication \cite{velivckovic2017graph}
& $0.79 \pm 0.07$ & $0.21 \pm 0.07$ & $0.70 \pm 0.07$ & $0.30 \pm 0.07$ & $0.83 \pm 0.06$ & $0.17 \pm 0.06$ & $0.71 \pm 0.06$ & $0.29 \pm 0.06$ \\
\midrule

Centralized Behavioral Cloning (BC) 
& $0.63 \pm 0.05$ & $0.37 \pm 0.05$ & $0.56 \pm 0.02$ & $0.44 \pm 0.02$ & -- & -- & -- & -- \\

Privileged CADRL \cite{everett2021collision}
& $1.0 \pm 0.00$ & $0.0 \pm 0.00$
& $0.90 \pm 0.00$ & $0.05 \pm 0.00$
& $0.98 \pm 0.00$ & $0.00 \pm 0.00$
& $0.95 \pm 0.00$ & $0.00 \pm 0.00$ \\
\bottomrule
\end{tabular}
}
\end{table*}

\paragraph{Quantitative comparison}
Table~\ref{tab:main_results} summarizes the performance of the proposed method in comparison with centralized policies with access to privileged information, decentralized policies without communication, and ablated variants of our framework. All methods (except CADRL) are trained on 4-agent environments and evaluated on both 4-agent and 8-agent settings. We report success rate and collision rate, averaged over 10 test scenarios and 4 random seeds. Across both agent scales, the proposed method consistently outperforms decentralized baselines and achieves performance close to that of privileged controllers. Notably, the centralized BC baseline, despite access to all agents' observations, lacks a shared global reference frame and thus struggles with relative coordination. In contrast, our learned communication (requiring no global frame at inference) induces a structured interaction representation that enables selective attention to relevant peers, effectively resolving coordination under partial observability. These results suggest that learned communication provides both a decentralization mechanism and a beneficial inductive bias for multi-agent coordination, with privileged training distilling interaction-relevant structure into lightweight, targeted signals.

\paragraph{Robustness to heterogeneous behaviors}
To further stress-test the learned policy, we evaluate scenarios in which a subset of two agents follow a non-cooperative policy that ignores others and prioritizes goal-reaching exclusively. Even under this partial policy mismatch, agents using the proposed communication-conditioned generative policy maintain a high success rate and avoid most collisions. This suggests that the learned policy does not rely on strong assumptions of fully cooperative behavior and exhibits robustness to partially adversarial or unmodeled agent dynamics, adapting effectively in many cases.

\subsection{Robustness to Communication Dropouts via Receding-Horizon Inference}
\label{sec:results_dropout}

\begin{figure}[b]
    \centering
    \includegraphics[width=\linewidth]{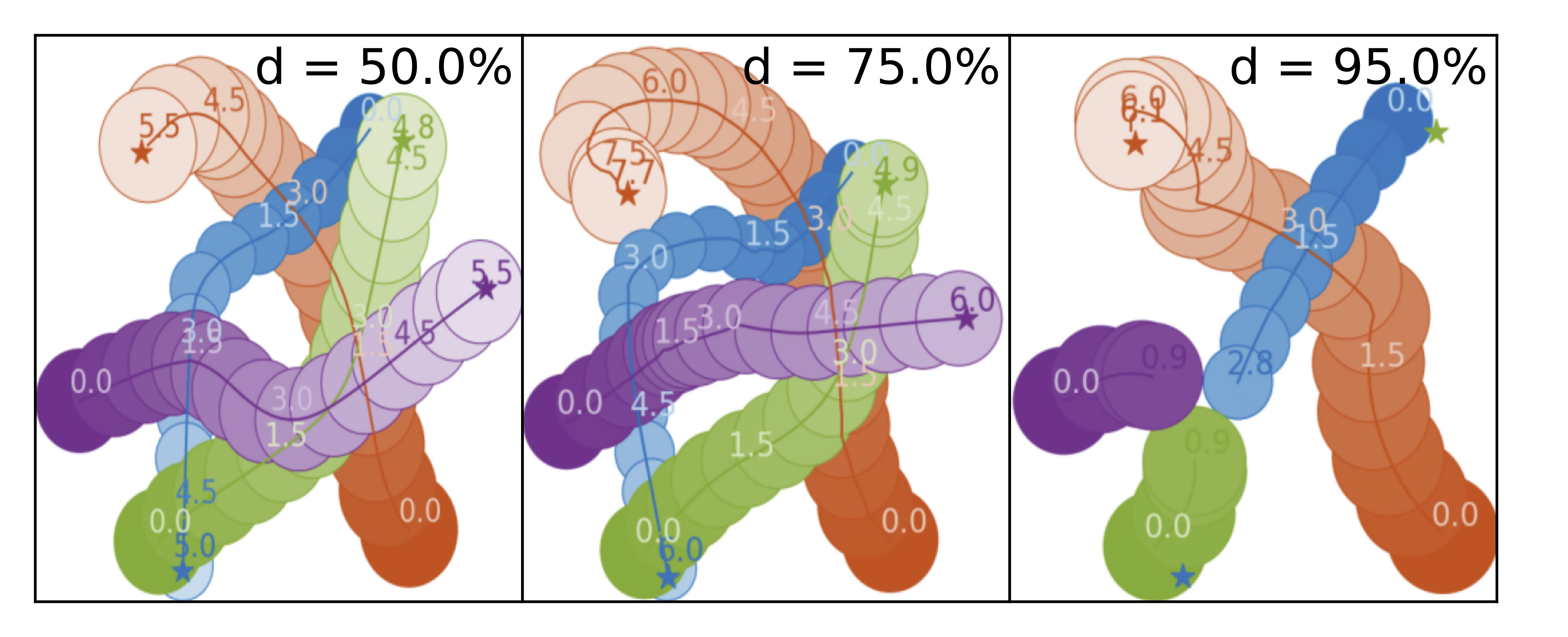}
    \vspace{-10pt}
    \caption{Robot trajectories under severe communication dropouts ($d=50\%,75\%,95\%$). Circles denote robot footprints, and stars denote goals.}
    \label{fig:dropout-results-figure}
\end{figure}

We evaluate the robustness of the proposed framework to intermittent or missing inter-agent communication, a common failure mode in real-world multi-robot systems due to bandwidth limits, packet loss, or hardware constraints. A central design objective of our receding-horizon generative policy is to degrade gracefully under such conditions, rather than relying on uninterrupted message exchange. To simulate communication failures, we introduce message dropouts during evaluation. For each episode, communication is disabled for a fraction $d \%$ of timesteps, during which agents receive no messages from other agents. Agents are not informed in advance when communication will be unavailable. All models are trained without dropout; robustness therefore arises purely from the inference-time design.

Figure~\ref{fig:dropout-results-figure} shows representative trajectories under increasing dropout rates. Even with communication disabled for $75\%$ of an episode, agents reliably reach their goals without collisions. At extreme dropout levels ($95\%$), performance degrades, with single agent completing the task. This behavior is expected given the deliberately limited local observations, which are insufficient to fully infer interaction-relevant properties of neighboring agents in the absence of communication. The observed robustness stems from the receding-horizon inference mechanism. When communication is available, agents refine short-horizon action sequences using communication-conditioned flow matching. When messages are unavailable, agents fall back to warm-started action plans inferred in previous timesteps, providing a temporally coherent default behavior. In addition, the policy can optionally perform a lightweight refinement step by perturbing the warm-started plan and denoising it conditioned on the latest local observation alone, enabling limited adaptation even without new messages.

\begin{figure}[!h]
    \centering
    \includegraphics[width=\linewidth]{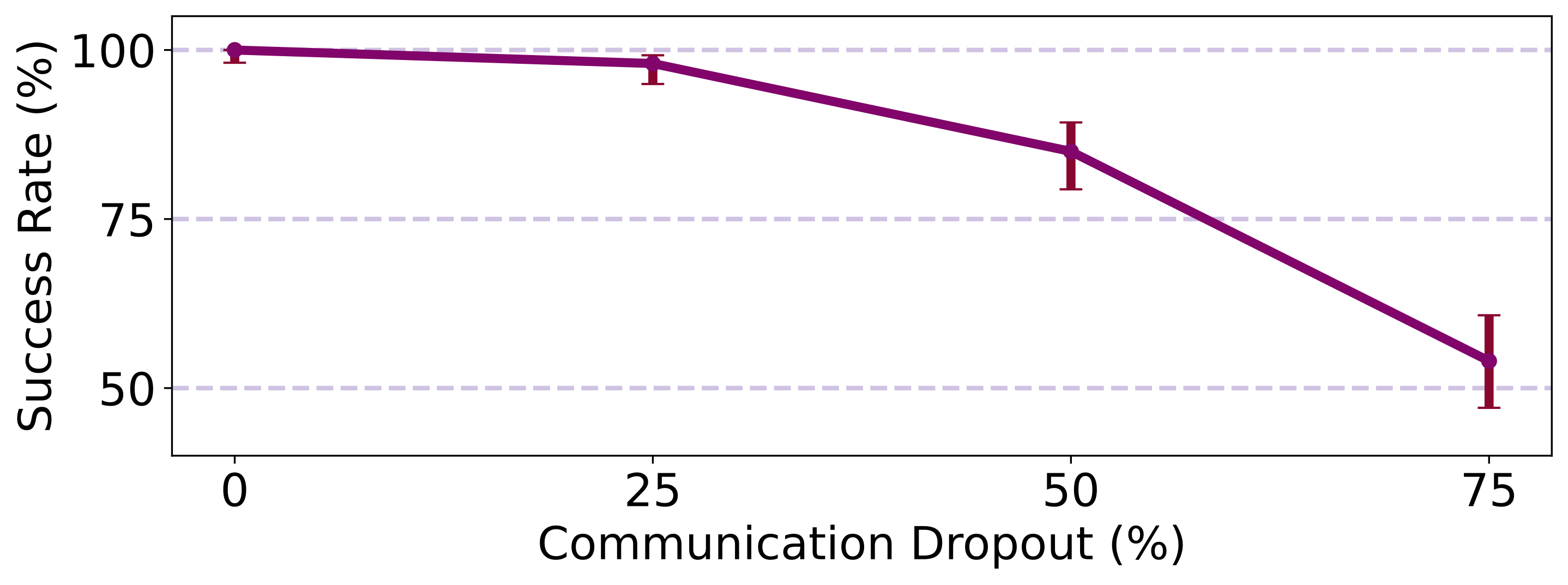}
    \vspace{-20pt}
    \caption{Success rate vs. communication dropout (50 runs per condition; 4 communicating agents per run; aggregated over 200 agent trials).}
    \label{fig:montecarlo_dropout}
\end{figure}

Figure~\ref{fig:montecarlo_dropout} reports success rates averaged over 50 randomized evaluation episodes (200 agent trials per condition). Performance degrades smoothly as communication dropout increases, demonstrating graceful degradation under moderate communication loss. Initial states, goal locations, and robot size are randomized across runs; the mean of such configurations is used for qualitative visualization in Figure~\ref{fig:dropout-results-figure}.

\subsection{Flexible Coordination via Flow Matching}
\label{sec:results_flexible_conditioning}

\begin{figure}[!h]
    \centering
    \includegraphics[width=0.9\linewidth]{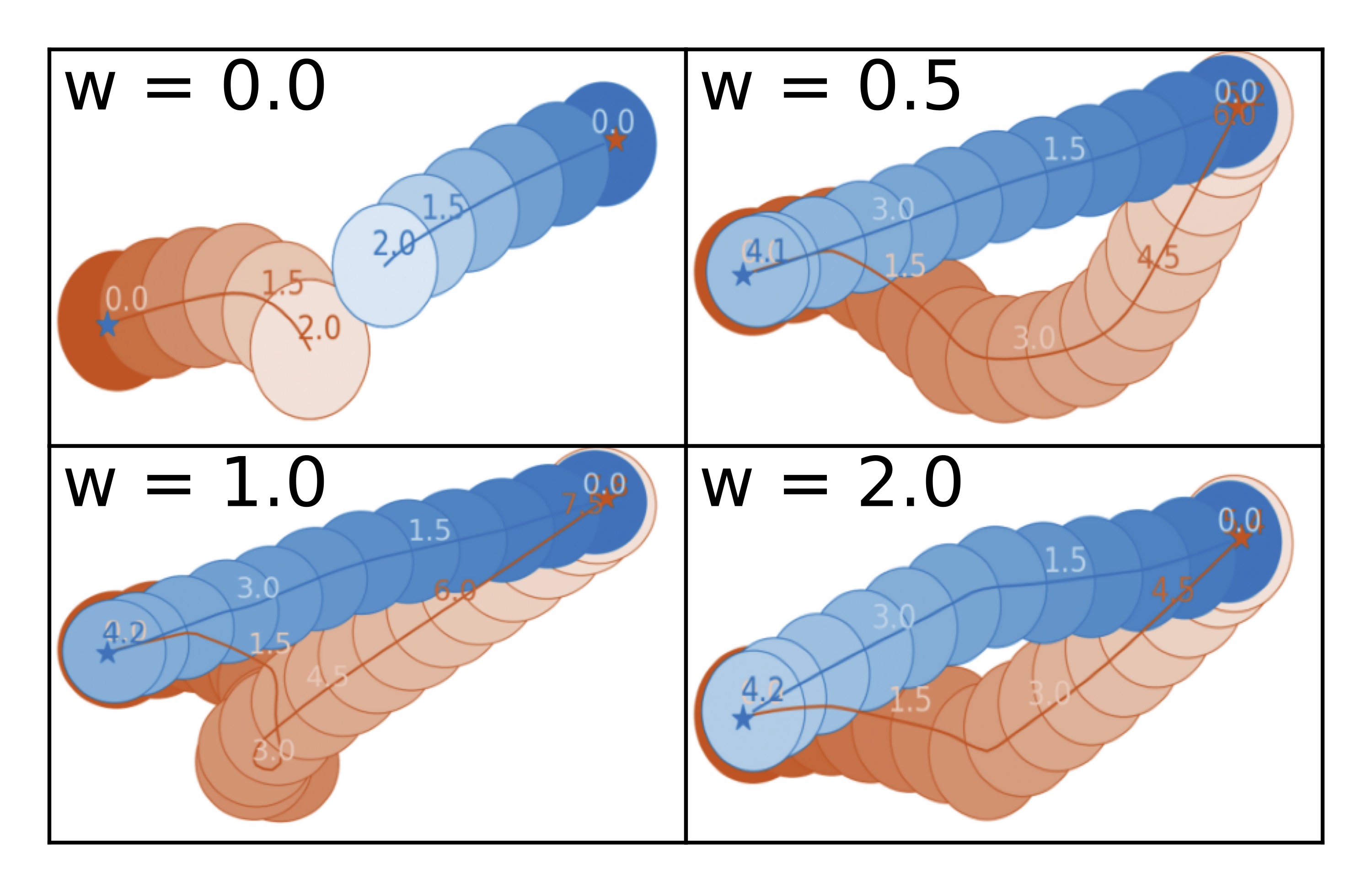}
    \vspace{-10pt}
    \caption{Trajectories generated by blending unconditioned and communication-conditioned velocity fields with guidance weight $w$. Increasing $w$ strengthens the influence of inter-agent messages, yielding a continuum from independent ($w=0.0$) to strongly coordinated ($w=2.0$) behavior. The policy is trained with four agents and evaluated zero-shot on a two-agent collision avoidance scenario.}
    \label{fig:cfg-results-figure}
\end{figure}

An emergent property of our communication-conditioned generative policy is its ability to smoothly interpolate between unconditioned and communication-conditioned action generation at inference time, in a manner reminiscent of classifier-free guidance (CFG), also recently used as policy improvement operator \cite{frans2025diffusion}. Our policy is trained to model both an unconditioned velocity field $u_\theta(o_i^{(t)}, \mathbf{A}_{\tau,i}^{(t)}, \tau, \varnothing)$, corresponding to a behavior based solely on local observations, and a communication-conditioned velocity field $u_\theta(o_i^{(t)}, \mathbf{A}_{\tau,i}^{(t)}, \tau, \bar{\mathbf{m}}_i^{(t)})$ that incorporates aggregated inter-agent messages. At inference, these outputs can be combined via a guidance formulation
\[
\hat{u}_\theta = (1-w)\,u_\theta^{\text{uncond}} + w\,u_\theta^{\text{cond}},
\]
where $w$ controls the influence of communication. Setting $w=0$ recovers purely local observation-driven, interaction-agnostic behavior, while $w=1$ yields fully communication-conditioned action generation; intermediate and extrapolated values induce a continuous spectrum of interaction behaviors. As illustrated in Fig.~\ref{fig:cfg-results-figure}, varying $w \in \{0.0,0.5,1.0,2.0\}$ produces progressively more coordinated trajectories in a two-agent collision avoidance scenario. Although flexible conditioning is not an explicit training objective, it emerges from jointly modeling conditioned and unconditioned flows and generalizes zero-shot when a policy trained with four agents is evaluated on two agents, underscoring the representational flexibility induced by the flow-matching design.

\subsection{Real-World Multi-Robot Navigation}
\label{sec:results_real_robot}

\begin{figure}[!h]
    \centering
    \includegraphics[width=0.9\linewidth]{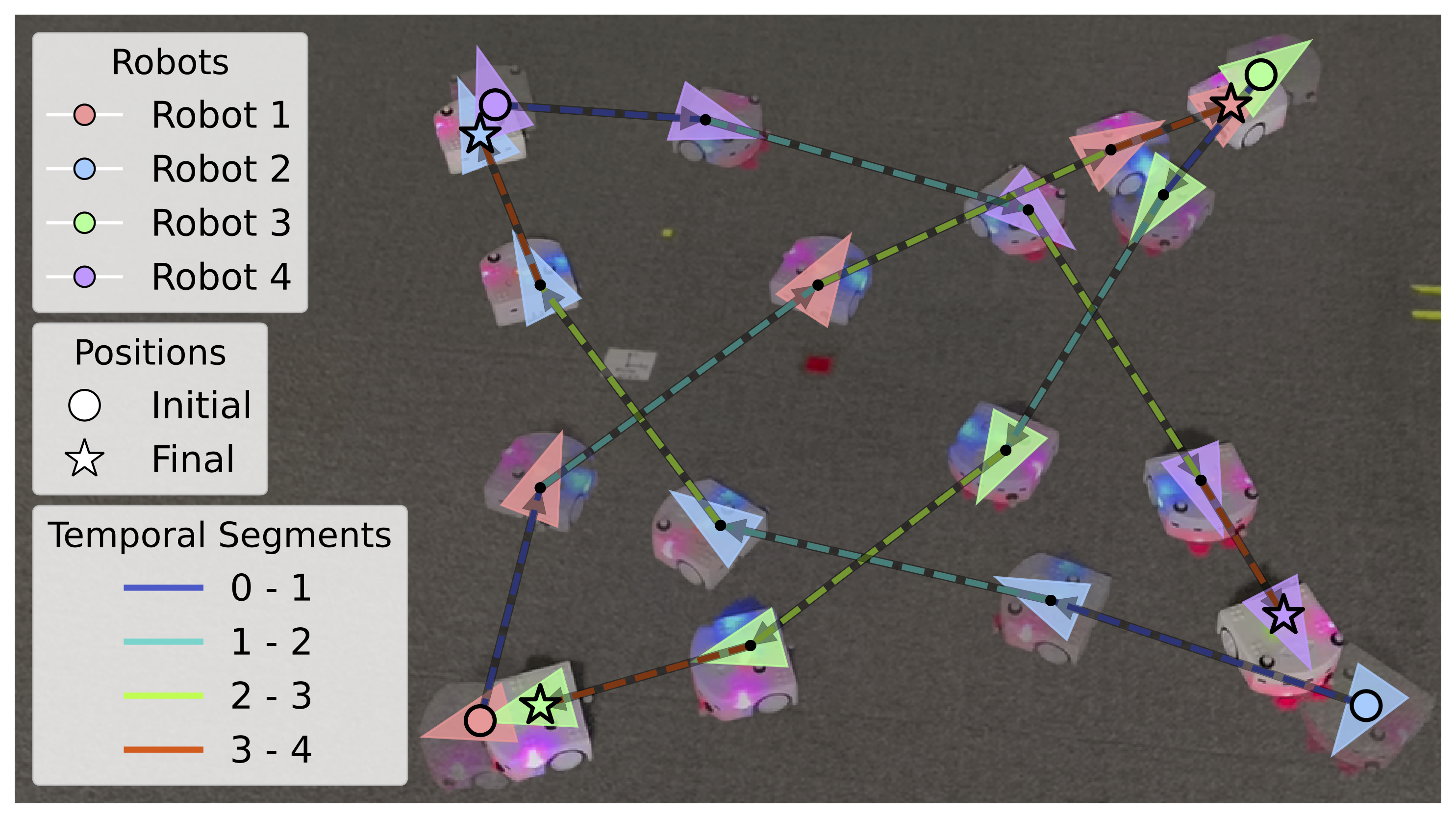}
    \vspace{-5pt}
    \caption{Zero-shot deployment on four Thymio robots swapping diagonal positions. Five keyframes are overlaid to illustrate temporal progression (increasing robot opacity indicates later times), with arrowed trajectory segments and colored triangular markers denoting agent indices and headings.}
    \label{fig:real-world-robot-exp}
\end{figure}

We next deploy the policy trained entirely in simulation directly on physical Thymio robots, without additional fine-tuning. Each robot uses local state estimates and executes the communication-conditioned policy in a fully decentralized manner. For robust transfer, instead of applying the raw predicted controls, we treat the predicted short-horizon action sequence as a local plan. We forward-propagate the controls using Eq. ~\eqref{unicycle_agent_dynamics} to obtain a look-ahead waypoint and track it with a simple proportional controller. This reduces sim-to-real transfer to waypoint tracking while leaving all multi-robot interaction and collision avoidance to the learned policy. No additional safety layers, privileged information or centralized coordination signals are used in this low-level controller. Figure~\ref{fig:real-world-robot-exp} shows four robots swapping diagonal positions, requiring tight mid-field interactions. The task is completed zero-shot, demonstrating transferable multi-agent coordination under real-world sensing and actuation noise.

\section{Conclusion}
We present a decentralized framework for communication-conditioned multi-agent collision avoidance based on flow-matching generative policies. By conditioning short-horizon action sequence generation on learned inter-agent messages, the proposed approach enables interaction-aware decision-making under partial observability without centralized control at execution time. Our formulation leverages offline demonstrations from a privileged expert while allowing communication to emerge end-to-end, resulting in a flexible inference mechanism that interpolates between independent and coordinated behavior. Combined with receding-horizon generation, this design yields smooth closed-loop behavior and graceful degradation under severe communication dropouts. Empirical results in dense multi-agent simulation and zero-shot real-robot deployment demonstrate that the proposed method achieves near-expert performance while retaining scalability and robustness. We believe this work highlights the promise of generative policies as a principled foundation for communication-aware multi-agent autonomy and opens avenues for extending flow-based decision-making to more complex interactive robotic systems.

\bibliographystyle{IEEEtran}
\bibliography{IEEEabrv,bib}

\end{document}